\documentclass[10pt,twocolumn,letterpaper]{article}

\usepackage{amsmath,amsfonts,bm}

\def\eqref#1{equation~\ref{#1}}

\def\1{\bm{1}}

\def\ve{{\bm{e}}}

\def\vv{{\bm{v}}}

\def\vx{{\bm{x}}}
\def\vy{{\bm{y}}}
\def\vz{{\bm{z}}}

\DeclareMathAlphabet{\mathsfit}{\encodingdefault}{\sfdefault}{m}{sl}
\SetMathAlphabet{\mathsfit}{bold}{\encodingdefault}{\sfdefault}{bx}{n}

\usepackage{cvpr}      

\usepackage[accsupp]{axessibility}
\usepackage{graphicx,adjustbox}
\usepackage{url}
\usepackage{amssymb}
\usepackage[linesnumbered,ruled,vlined]{algorithm2e}
\SetAlFnt{\small}
\SetKwComment{Comment}{/* }{ */}
\SetKwInput{kwInput}{Input}
\SetKwInput{kwOutput}{Output}
\RestyleAlgo{ruled}
\usepackage{amsmath}
\usepackage{array}
\usepackage{multirow}
\usepackage{booktabs}
\usepackage[table]{xcolor}
\usepackage[normalem]{ulem}
\useunder{\uline}{\ul}{}
\usepackage{float, wrapfig}
\usepackage{resizegather}
\usepackage{enumitem}
\def\code#1{\texttt{#1}}

\usepackage[font=small,skip=0pt]{caption}
\usepackage{amssymb}

\usepackage[pagebackref,breaklinks,colorlinks]{hyperref}
\usepackage[capitalize]{cleveref}
\crefname{section}{Sec.}{Secs.}
\Crefname{section}{Section}{Sections}
\Crefname{table}{Table}{Tables}
\crefname{table}{Tab.}{Tabs.}

\def\cvprPaperID{14016} 
\def\confName{CVPR}
\def\confYear{2024}
\begin{document}

\title{NAYER: Noisy Layer Data Generation for Efficient and Effective Data-free Knowledge Distillation}

\author{Minh-Tuan Tran$^1$, Trung Le$^1$, Xuan-May Le$^2$, Mehrtash Harandi$^1$, Quan Hung Tran$^3$,\\ Dinh Phung$^{1,4}$\\\
$^1$Monash University,  $^2$University of Melbourne, $^3$ServiceNow, $^4$VinAI Research\\
{\tt\small \{tuan.tran7,trunglm,mehrtash.harandi,dinh.phung\}@monash.edu} \\ \tt\small xuanmay.le@student.unimelb.edu.au, hungquan.tran@servicenow.com}
\maketitle

\begin{abstract}

Data-Free Knowledge Distillation (DFKD) has made significant recent strides by transferring knowledge from a teacher neural network to a student neural network without accessing the original data. Nonetheless, existing approaches encounter a significant challenge when attempting to generate samples from random noise inputs, which inherently lack meaningful information. Consequently, these models struggle to effectively map this noise to the ground-truth sample distribution, resulting in prolonging training times and low-quality outputs. In this paper, we propose a novel Noisy Layer Generation method (NAYER) which relocates the random source from the input to a noisy layer and utilizes the meaningful constant label-text embedding (LTE) as the input.  LTE is generated by using the language model once, and then it is stored in memory for all subsequent training processes. The significance of LTE lies in its ability to contain substantial meaningful inter-class information, enabling the generation of high-quality samples with only a few training steps. Simultaneously, the noisy layer plays a key role in addressing the issue of diversity in sample generation by preventing the model from overemphasizing the constrained label information. By reinitializing the noisy layer in each iteration, we aim to facilitate the generation of diverse samples while still retaining the method's efficiency, thanks to the ease of learning provided by LTE. Experiments carried out on multiple datasets demonstrate that our NAYER not only outperforms the state-of-the-art methods but also achieves speeds 5 to 15 times faster than previous approaches. The code is available at \url{https://github.com/tmtuan1307/nayer}.
\end{abstract}

\section{Introduction}

\begin{figure}
\includegraphics[width=1\linewidth]{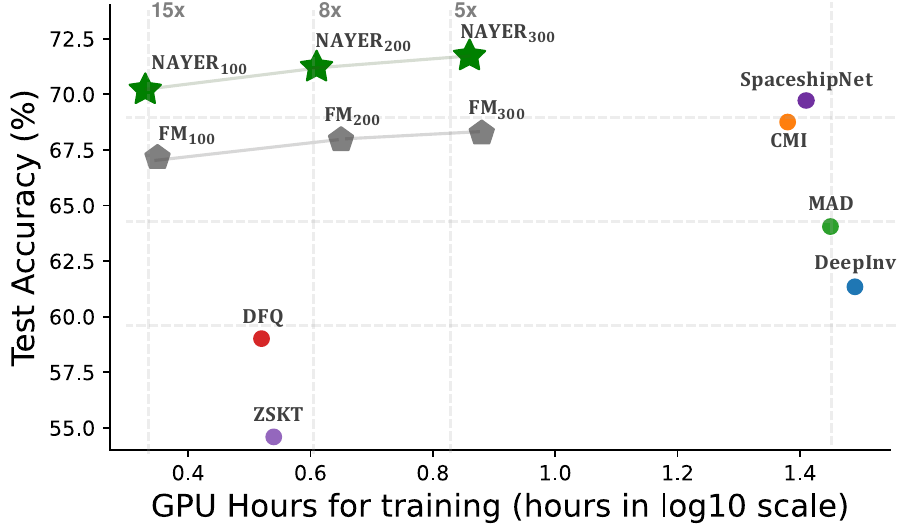} 
\caption{Accuracy of student models and GPU hours of training time on CIFAR-100 dataset. All variants of our method NAYER not only attains the highest accuracies across but also accelerates the training process by 5 to 15 times compared to DeepInv \cite{adi}.}
\label{fig:time_vs_acc}
\end{figure}

Knowledge distillation (KD) \cite{kd,kd1,kd2,kd4,kd5} aims to train a student model capable of emulating the capabilities of a pre-trained teacher model. Over the past decade, KD has been explored across diverse domains, including image recognition \cite{kd3, uda}, software engineering \cite{se}, and natural language processing \cite{nlpkd}. Conventional KD methods generally assume that the student model has access to all or part of the teacher's training data. However, real-world applications often impose constraints on accessing the original training data. This issue becomes particularly relevant in cases of privacy-sensitive medical data, which may contain personal information or data considered proprietary by vendors. Consequently, in such contexts, conventional KD methods no longer suffice to address the challenges posed.

\begin{figure*}[t]
\begin{center}
\includegraphics[width=0.9\linewidth]{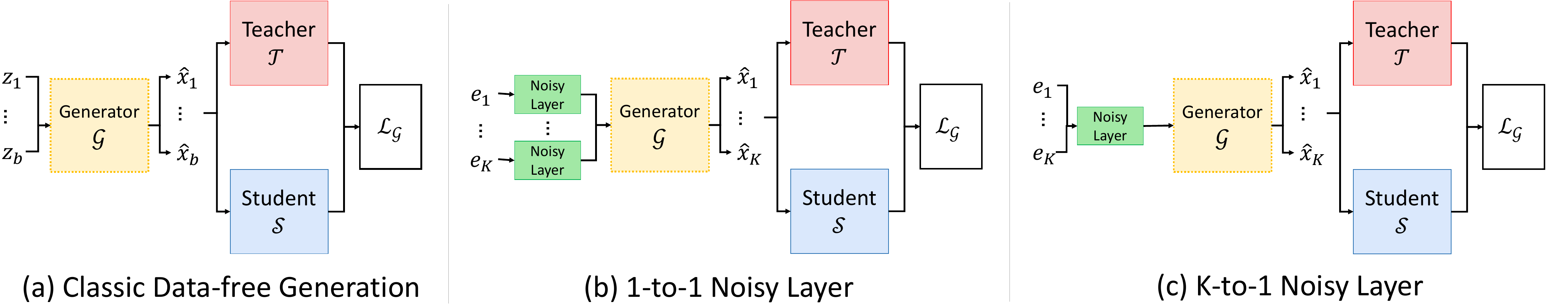}
\end{center}
\caption{Data Generation Strategies: (a) Classic method which optimizes random noise (z); (b) Using one noisy layer for generating one synthetic image from the label-text embedding ($\ve_{\vy}$); (c) Using one noisy layer to generate multiple synthetic images.}
\label{fig:m1}
\end{figure*}

Data-Free Knowledge Distillation (DFKD) \cite{adi,cmi,spshnet,lander,kakr,mad,dfkd_gr} has recently seen significant advancements as an alternative method. Its core principle involves transferring knowledge from a teacher neural network ($\mathcal{T}$) to a student neural network ($\mathcal{S}$) by generating synthetic data instead of accessing the original training data. The synthetic data enable adversarial training of the generator and student \cite{zskd,zskt}. In this setup, the student seeks to match the teacher's predictions on synthetic data, while the generator aims to create samples that maximize the discrepancy between the student's and teacher's predictions (Figure~\ref{fig:m1}a). 

Due to its reliance on synthetic samples, the need for an effective and efficient data-free generation technique becomes imperative. A major limitation of current DKFD methods is that they merely generate synthetic samples from random noise, neglecting to incorporate supportive and semantic information \cite{predfkd,fastdfkd,spshnet,kakr}. This limitation in turn incurs the generation of low-quality data or excessive time requirements for training the generator, rendering them unsuitable for large-scale tasks. Notably, almost state-of-the-art (SOTA) DFKD methods  do not report results on large-scale ImageNet due to the significant training time involved. Even with smaller datasets such as CIFAR-100 (see Figure~\ref{fig:time_vs_acc}), SOTA DFKD methods such as CMI \cite{cmi}, MAD \cite{mad}, or DeepInv still demand approximately 25 to 30 hours of training while struggling to achieve high accuracy. This emphasizes the pressing need for more efficient and effective DFKD techniques.

{\color{black} To address mentioned problem, we introduce a simple yet effective DFKD method called  \underline{N}oisy l\underline{AYER} generation (NAYER). Our approach relocates the source of randomness from the input to the noisy layer and utilizes the meaningful label-text embedding (LTE) generated by a pre-trained language model (LM) \cite{sbert, clip, doc2vec} as the input. In this context, LTE plays a crucial role in accelerating the training process due to its ability to encapsulate useful interclass information. Note that, there is a common observation that the text with similar meanings tend to exhibit closer embedding proximity to one another \cite{w2v,v2,v3}. For instance, the text embedding of sentence \textit{"A class of a dog"} and \textit{"A class of a cat"} is always closer compared to \textit{"A class of a car"}. Consequently, by using LTE as input, our approach can proficiently generate high-quality samples that closely mimic the distributions of their respective classes with only a few training steps. It is important to note that our method only queries the LTE from the LM once. This LTE is then stored in memory for subsequent processing, and we do not use the language model in the training process.

However, when utilizing a constant LTE as the input, we empirically observed that the generator consistently produces a set of similar data lacking diversity in every iteration. Our solution addresses this issue by introducing the layer-level random source by adding a noisy layer (NL) to learn the constant label information.  This involves incorporating a random NL to function as an intermediary between the generator and LTE, which prevents the generator from relying solely on unchanging label information (Figure~\ref{fig:m1}b). The source of randomness now comes from the random reinitialization of the NL for each iteration. Through this mechanism, we aim to effectively mitigate the risk of overemphasizing label information, thus enhancing the diversity of synthesized images. Furthermore, thanks to the inherent ease of learning label-text embeddings, regardless of how it is initialized, jointly training the NL with generator can consistently generate high-quality samples in just a few steps, thereby maintaining the method's efficiency. Additionally, we propose leveraging a single NL to generate multiple samples (e.g., 100 images across 100 classes of CIFAR-100) (Figure~\ref{fig:m1}c). 
This strategy reduces the number of training parameters and enhances diversity by leveraging multiple gradient sources from various classes.

Our major contributions are summarized as follows:
\begin{itemize}[noitemsep,nolistsep]
    \item We propose NAYER, a simple yet effective DFKD method based on LTE and a noisy layer, providing the fast training with high classification performance.
    \item We introduce a K-to-1 noisy layer, which utilizes only a single noisy layer to generate multiple samples.
    \item Experiments on CIFAR100, TinyImageNet and ImageNet, demonstrate that our method ourperforms SOTA algorithms in both accuracy and training time. Specifically, our methods achieve speeds that are 5 to even 15 times faster while also attaining higher accuracies compared to previous methods (Figure~\ref{fig:time_vs_acc}).
    \item To the best of our knowledge, we are the first to introduce the use of LTE and the concept of layer-level random source for DFKD.
\end{itemize}
}
\section{Related Work}
\noindent
\textbf{Data-Free Knowledge Distillation.} DFKD methods \cite{adi,cmi,spshnet,mad,kakr} generate synthetic images to transfer knowledge from a pre-trained teacher model to a student model. These data are used to jointly train the generator and the student in an adversarial manner \cite{zskt}. In this adversarial learning scheme, the student aims to make predictions close to the teacher’s on synthetic data, while the generator endeavors to create samples that align with the teacher's confidence but also maximize the mismatch between the student’s and the teacher’s predictions. This adversarial game enables an rapid exploration of synthetic distributions useful for knowledge transfer between the teacher and the student. 

\noindent
\textbf{Data-Free Generation.} As the central principle of DFKD revolves around synthetic samples, the data-free generation technique plays a pivotal role. \cite{adi} proposes the image-optimized method which attempts to optimize the random noise images using teacher network batch normalization statistics. Sample-optimized methods \cite{cmi,spshnet} focus on optimizing random noise over numerous training steps to produce synthetic images in case-by-case strategy. In contrast, generator-optimized methods \cite{mad,kakr,predfkd} attempt to ensure that the generator has the capacity to comprehensively encompass the entire distribution of the original data. In the other words, regardless of the input random noise, these methods aim to consistently yield high-quality samples for training the student model. This approach often prolongs the training process and may not consistently produce high-quality samples, particularly when diverse noises are employed during both the sampling and training phases.  Furthermore, the main problem in existing data-free generation is the use of random noise input without any meaningful information, leading to generate the low-quality samples and prolonged training times for the generator. \cite{fastdfkd} introduced FM, a method incorporating a meta generator to accelerate the DFKD process significantly. However, this acceleration comes at the cost of a noticeable trade-off in accuracy.

\noindent
\textbf{Synthetizing Samples from Label Information}. Drawing inspiration from the success of incorporating label information in adversarial frameworks like Conditional GAN \cite{cgan,cgan2,cgan3}, several DFKD methods have adopted strategies to generate images guided by labels. In these approaches, a common practice involves fusing random noise ($\vz$) with a learnable embedding ($\ve_\vy$) of the one-hot label vector, which is used as input for the model \cite{cgdfkd1, cgdfkd2, mad}. This combination enhances control over the resulting class-specific synthetic images. However, despite the potential of label information, its application has yielded only minor improvements. This can be attributed to two key factors. Firstly, the one-hot vector introduces sparse information that merely distinguishes labels uniformly, failing to capture the nuanced relationships between different classes. Consequently, the model struggles to generate images that align closely with ground-truth distributions. Secondly, there exists a challenge in balancing the generated images' quality and diversity when incorporating label information. This can inadvertently lead to an overemphasis on label-related details, potentially overshadowing the crucial contribution of random noise, which is necessary for generating a diverse range of samples

\section{Proposed Method}
\subsection{Problem Formulation}


Consider a training dataset $D =\{ (\vx_i, \vy_i)\}_{i=1}^m $ with $\vx_i \in \mathbb{R}^{c \times h \times w}$ and $\vy_i \in \{1,2,\cdots,K\}$, where the pair $(\vx_i, \vy_i)$ represents a training sample and its corresponding label, respectively. Let $\mathcal{T}=\mathcal{T}_{\theta_\mathcal{T}}$ be a pre-trained teacher network on $D$. The objective of DFKD is to train a student network $\mathcal{S}=\mathcal{S}_{\theta_\mathcal{S}}$ to emulate $\mathcal{T}$'s performance, all without needing access to the original dataset $D$. 

To achieve this, we employ the lightweight generator $\mathcal{G}_{\theta_\mathcal{G}}$ to generate synthetic images and subsequently use them to train a student network $\mathcal{S}$. Specifically, in contrast to existing DFKD methods \cite{spshnet, mad, kakr, adi, cmi, fastdfkd}, our approach utilizes a meaningful constant label-text embedding (LTE) as the input for $\mathcal{G}$ instead of random noise. Due to LTE's capability to encapsulate valuable interclass information, this accelerates the generation process, expediting the training time (Section \ref{sec:lte}). Following that, we propose the use of a layer-level random source (Noisy Layer) to better adapt with LTE for generating diverse synthetics (Section \ref{sec:nl}). Finally, the synthetic images are employed for the joint training of the generator and student in an adversarial manner to enhance knowledge transfer (Section \ref{sec:g}).

\subsection{Label-Text Embedding as Generator's Input}
\label{sec:lte}

\begin{figure*}[t]
\begin{center}
\includegraphics[width=0.9\linewidth]{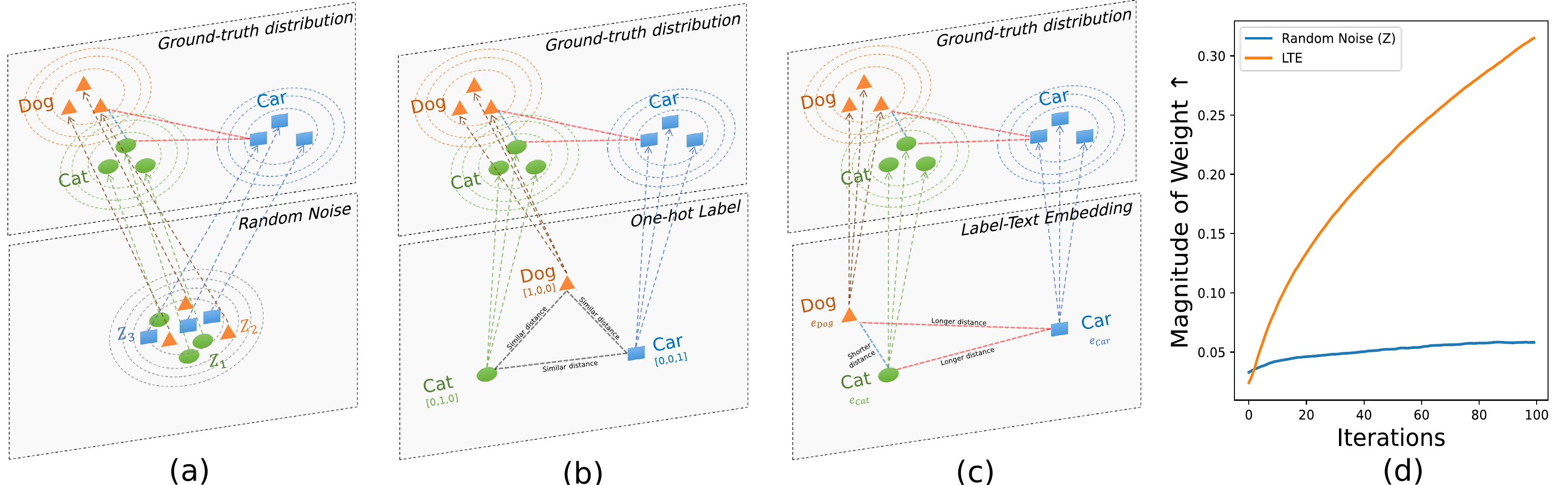}
\end{center}
\caption{(a) Random noise for data generation. (b) One-hot labels only uniformly distinguish labels, lacking inter-class relationships. In contrast, (c) LTE captures inter-class connections, bringing similar classes closer in the embedding space. This proximity enhances the similarity between the input and ground-truth sample distributions, thereby allowing the model to more easily mimic the ground-truth distribution and accelerating the learning process. (d) The averaging magnitude of weight used to learn LTE is much larger than those for random noise, highlighting the model's negative focus on label information while ignoring random noise. }
\label{fig:motivation}
\end{figure*}

The main limitation of existing DFKD methods is synthetize data from random noise, which have no supportive and semantic information. Therefore, they usually generate very low-quality data \cite{fastdfkd} or require a excessive training time for high quality image generation \cite{mad,kakr,spshnet,cmi,adi}. There are also several methods use to one-hot vector of classes as the additional input to resemble the conditional generator, however its application has yielded only minor improvements. The main reason is the one-hot vector (OH) introduces sparse information which make a generator hard to learn about it. Furthermore, OH merely distinguishes labels uniformly, failing to capture the nuanced relationships between different classes. 

To address this problem, we are the first to propose the use of label-text embeddings for DFKD by employing them as an input for the generator. LTE, as a dense vector with richer information, facilitates an easier learning process for the model. Additionally, LTE capitalizes on the tendency for text with similar meanings to exhibit proximity in their embeddings \cite{doc2vec}. Figure~\ref{fig:motivation}a-c visually represents the LTE, highlighting their superior capacity to depict the relationship between the `Dog' and `Cat' classes. This is evident in their closer proximity (shorter distance) when compared to the `Car' class. This characteristic of LTE contributes to making the input distribution (representing labels)  and ground-truth distribution  (representing actual data) more similar. As a result, it facilitates the model's mapping between these two distributions, accelerating the learning process and generating high-quality images. 

\noindent
\textbf{Prompt Engineering.} Given the list of all classes $\vy = [\vy_1,\cdots,\vy_K]$, their label text $Y_{\vy} = [Y_{\vy_1},\cdots,Y_{\vy_K}]$ is generated by using a manually designed prompt template such as \code{"a class of a \{class\_name\}"}. Then, the label-text prompt is then embedded using a pre-trained text encoder $\mathcal{C}$ as follows: 
\begin{equation}
    \ve_{\vy} = \mathcal{C}(Y_{\vy})\;.
    \label{eq:ey}
\end{equation} 

\noindent
\textbf{LTE Pool.} Importantly, the embedding $\ve_{\vy}$ is generated once and then stored in the LTE pool $\mathcal{P}$, remaining fixed throughout the entire training process. The text encoder $\mathcal{C}$ is not utilized during the training process. In training phases, with a batch of pseudo-labels $\hat{\vy}$, we retrieve their corresponding LTEs from $\ve_{\hat{\vy}}\sim\mathcal{P}$ and employ these LTEs as inputs for the generator. This eliminates the reliance on random noise for synthetic image generation. 
\begin{equation}
    \hat{\vx} = \mathcal{G}(\ve_{\hat{\vy}})\;.
    \label{eq:genxnew}
\end{equation}

We conducted an ablation study to analyze the impact of different prompt engineering template and language model (LM) for generating LTEs in Section \ref{sec:ana}. 


Thanks to the informative content embedded in LTE, our approach can efficiently produce high-quality samples with minimal computational steps. We have also conducted an empirical study to substantiate this claim, as illustrated in Table \ref{tab:ab_lte_nayer}. The results of this study highlight that LTE significantly accelerates convergence in terms of Cross-Entropy (CE) Loss and yields higher-quality images (as measured by the Inception Score or IS score) compared to random noise and one-hot vectors. This acceleration empowers our method to achieve convergence with a considerably smaller training steps for generator (30 steps for CIFAR10 and 40 steps for CIFAR100), compared to the 2,000 steps required by DeepInv or the 500 steps of CMI, all while maintaining superior accuracy (as detailed in Table~\ref{tab:sota}).

\subsection{Generating Diverse Samples with Noisy Layer}
\label{sec:nl}

While leveraging label information provides advantages for data generation, the synthetic images are less diverse due to the absence of a random source. Two common solutions involve concatenating random noise $\vz$ and $\ve_\vy$ or using their sum as the generator input. However, both approaches have limitations. Concatenation raises the risk of overemphasizing label-related information, as evidenced by significantly larger weight magnitudes for learning LTE compared to random noise (Figure~\ref{fig:motivation}d), which can be seen as the significance of these weights \cite{lottery}. Using the sum of $\vv = \ve_{\hat{\vy}} + \beta\vz$ faces challenges: a low $\beta$ results in an insufficient random source for diverse sampling, and a high $\beta$ may overshadow LTE features, leading to a reliance on random noise $\vz$. This challenge is also observed in some existing methods \cite{cgdfkd1, mad}, where the application of the sum of noise and label information provides minimal improvement compared to an unconditional generator. 

To effectively introduce randomness to LTE, we propose the concept of a layer-level random source with the Noisy Layer. The source of randomness now stems from the random reinitialization of the NL during each iteration. With each different initialization, the NL learns LTE in a distinct way, successfully mitigating the risk of a negative bias towards LTE. Unlike existing sources of randomness, the design of NL provides a larger random parameter to enhance the diversity of the synthesized images. Furthermore, due to the straightforward training of LTE, regardless of its initialization, the joint training of the noisy layer and the generator consistently yields high-quality samples within a few iterations, thus preserving the method's efficiency.

\begin{figure*}[t]
\begin{center}
\includegraphics[width=0.95\linewidth]{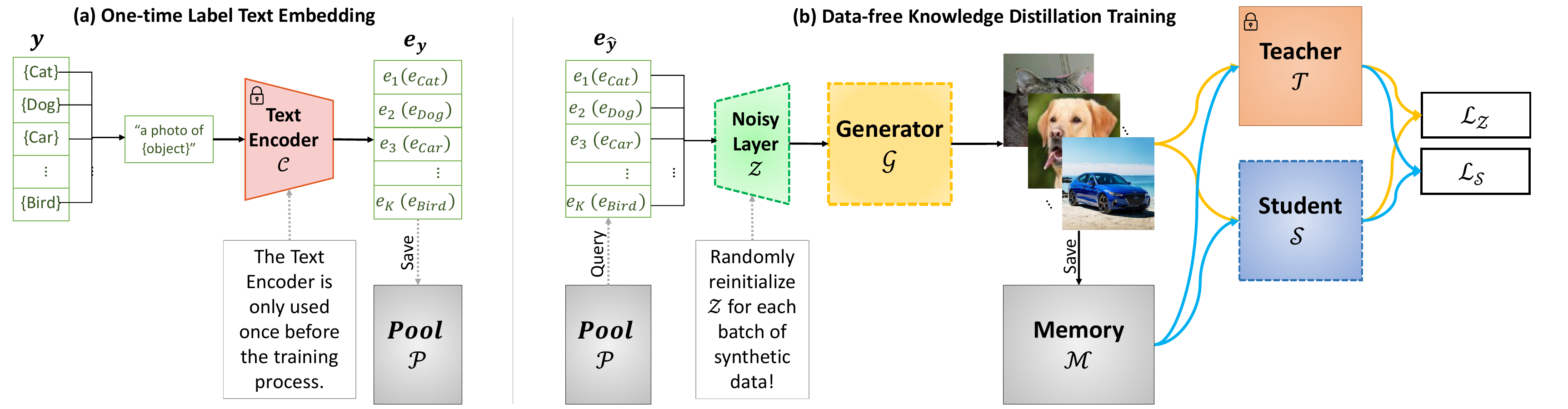}
\end{center}
\caption{General Architecture of Noisy Layer Generation for Data-free Knowledge Distillation: NAYER initially employs the text encoder to generate the LTEs, which are then stored in the memory pool for model training. In each training batch, the LTEs serve as input for the noisy layer $\mathcal{Z}$ and generator $\mathcal{G}$ to produce synthetic images. Finally, these images are used for the joint training of the generator, noisy layer, and student network using Eq. \ref{eq:lz} and Eq. \ref{eq:ls_m}.}
\label{fig:arch}
\end{figure*}

\noindent
\textbf{Noisy Layer Architecture.} We design the NL $\mathcal{Z}_{\theta_{\mathcal{Z}}}$ as a combination of a \code{BatchNorm} layer and a single \code{Linear} layer. The input size of the \code{Linear} layer matches the embedding size of the text encoder ($e$), and the output size corresponds to the noise dimension ($r$). Typically, this output size is set to 1,000, following to \cite{kakr, mad, spshnet}. The simplicity of the single \code{Linear} layer is crucial for expediting the generation process. It converges rapidly without requiring an excessive number of steps, yet its size remains sufficiently large to provide an sufficient random source for the generator. Additionally, a \code{BatchNorm} module plays a role in increasing the distance between LTEs (from averaging 0.015 to 0.45 using L2 distance), helping the model discriminate these LTEs easier and thereby speeding up the training process. Furthermore, with a different batch of $\hat{\vy}$, the output of \code{BatchNorm} can vary, introducing a slight additional randomness for the generator. The ablation study analyzing the impact of different architectures of NL can be found in the \textbf{Appendix}.

Given LTEs $\ve_{\hat{\vy}}$, we feed these $\ve_{\hat{\vy}}$ into the noisy layer $\mathcal{Z}$. Then, the output of the noisy layer is fed into the generator $\mathcal{G}$ to produce the batch of synthetic images $\hat{\vx}$:
\begin{align}
    \mathcal{Z}(\ve_{\hat{\vy}}) &= \code{Linear}(\code{BatchNorm}(\ve_{\hat{\vy}}))\;.\\
    \hat{\vx} &= \mathcal{G}(\mathcal{Z}(\ve_{\hat{\vy}}))\;.
    \label{eq:v}
\end{align}
\noindent
\textbf{K-to-1 Noisy Layer.} In the existing approach, a separate random source is created for each instance, as similar inputs generate similar samples. In contrast, we propose employing a single noisy layer to learn from all available classes (K-to-1) by inputting $\ve_{\hat{\vy}}$ with $\hat{\vy}= 1,\dots,K$ to a single noisy layer $\mathcal{Z}$. This design enables the noisy layer to generate multiple samples simultaneously, such as a maximum of 100 for CIFAR100 or 10 for CIFAR10, thus reducing a parameter size and efficiently expediting training. The underlying  idea revolves around the fact that each class has distinct LTEs. Thus, by supplying different inputs of $\ve_{\hat{\vy}}$ from K classes, the noisy layer can still generate diverse images. Furthermore, we also empirically observe that using a single noisy layer to synthesize a batch of images (K-to-1) enriches generator diversity, ensuring both fast convergence and high-quality sample generation. This enhancement can be attributed to the use of multiple gradient sources from diverse classes, which can further enriches the diversity of the noisy layer's output.

\subsection{Generator and Student Updating}
\label{sec:g}
\begin{algorithm}[t]
\caption{NAYER}
\label{alg:nldfkd}
\kwInput{pre-trained teacher $\mathcal{T}_{\theta_\mathcal{T}}$, student $\mathcal{S}_{\theta_\mathcal{S}}$, generator $\mathcal{G}_{\theta_\mathcal{G}}$, text encoder $\mathcal{C}_{\theta_\mathcal{C}}$, list of labels ${\vy}$ and list of text of these labels $Y_{{\vy}}$\; }
\kwOutput{An optimized student  $\mathcal{S}_{\theta_\mathcal{S}}$}
Initializing $\mathcal{P} = \{\}, \mathcal{M} = \{\}$\;
Store all embeddings $\ve_{{\vy}} = \mathcal{C}(Y_{{\vy}})$ into $\mathcal{P}$\;
\For{$\mathcal{E}$ \textit{epochs}}{
    \For{$I$ \textit{iterations}}{
        Randomly reinitializing noisy layers $\mathcal{Z}_{\theta_{\mathcal{Z}}}$ and pseudo label $\hat{\vy}$ for each iteration\;
        Query $\ve_{\hat{\vy}} \sim \mathcal{P}$\;
        \For{$g$ \textit{steps}}{
            $\hat{\vx} \gets \mathcal{G}(\mathcal{Z}(\ve_{\hat{\vy}}))$\;
            $\mathcal{L}_{\mathcal{Z}} \gets \alpha_{cls}\mathcal{L}_\text{CE}(\mathcal{T}(\hat{\vx}), \hat{\vy})-\alpha_{adv}\mathcal{L}_\text{KL}(\mathcal{T}(\hat{\vx}),\mathcal{S}(\hat{\vx}))+\alpha_{bn}\mathcal{L}_\text{BN}(\mathcal{T}(\hat{\vx}))$\;
            Update $\theta_{\mathcal{G}}, \theta_{\mathcal{Z}}$ by minimizing $\mathcal{L}_{\mathcal{Z}}$;
        }
        $\mathcal{M} \gets \mathcal{M} \cup \hat{\vx}$\;
    }
    \For{$S$ \textit{iterations}}{
    $\hat{\vx} \sim \mathcal{M}$\;
    Update $\theta_{\mathcal{S}}$ by minimizing $\mathcal{L}_\mathcal{S} \gets \mathcal{L}_\text{KL}(\mathcal{T}(\hat{\vx}),\mathcal{S}(\hat{\vx}))$;
    }
}
\end{algorithm}

To make it easier to follow, we provide the architecture of NAYER in Figure \ref{fig:arch} and the detailed pseudocode in Algorithm \ref{alg:nldfkd}, wherein NAYER initially embeds all label text using a text encoder. Subsequently, our method undergoes training for $\mathcal{E}$ epochs. Within each training epoch, NAYER consists of two distinct phases. The first phase involves training the generator. In each iteration $I$, as described in Algorithm \ref{alg:nldfkd}, the noisy layer $\mathcal{Z}$ is reinitialized (line 5) before being utilized to learn the LTE. The generator and the noisy layer are then trained through $g$ steps using Eq.~(\ref{eq:lz}) to optimize their performance (line 10).
\begin{multline}
    \min\limits_{\theta_\mathcal{G},\theta_\mathcal{Z}}\mathcal{L}_\mathcal{Z} \triangleq \mathbb{E}_{\hat{\vx} \sim \mathcal{G}(\mathcal{Z}(\ve_{\hat{\vy}}))}\Big[\alpha_{cls}\mathcal{L}_\text{CE}(\mathcal{T}(\hat{\vx}), \hat{\vy}) \\-\alpha_{adv}\mathcal{L}_\text{KL}(\mathcal{T}(\hat{\vx}),\mathcal{S}(\hat{\vx}))+\alpha_{bn}\mathcal{L}_\text{BN}(\mathcal{T}(\hat{\vx}))\Big]\;.
    \label{eq:lz}
\end{multline}
Within this context, $\mathcal{L}_\text{CE}$ represents the Cross-Entropy loss term, serving the purpose of training the student on images residing within the high-confidence region of the teacher's knowledge. Conversely, the negative $\mathcal{L}_\text{KL}$ term facilitates the exploration of synthetic distributions, boosting effective knowledge transfer between the teacher and the student. In other words, the student network takes on a role as a discriminator in GANs, ensuring the generator is geared towards producing images that the teacher has mastered, yet the student network has not previously learned. This approach facilitates the focused development of the student's understanding in areas where it lags behind the teacher, enhancing the overall knowledge transfer process. We also use batch norm regularization ($\mathcal{L}_\text{BN}$) \cite{adi,fastdfkd}, a commonly used loss in DFKD, to constrain the mean and variance of the feature at the \code{BatchNorm} layer to be consistent with the running-mean and running-variance of the same layer.

The second phase involves training the student networks. During this phase, all the generated samples are stored in the memory module $\mathcal{M}$ to mitigate the risk of forgetting (line 10), following a similar approach as outlined in \cite{fastdfkd}. Ultimately, the student model is trained by Eq.~(\ref{eq:ls_m}) over $S$ iterations, utilizing the samples from $\mathcal{M}$ (lines 13 and 14).
\begin{equation}
    \min\limits_{\theta_\mathcal{S}}\mathcal{L}_\mathcal{S} \triangleq \mathbb{E}_{\hat{\vx} \sim\mathcal{M}}\Big[\mathcal{L}_\text{KL}(\mathcal{T}_{\theta_\mathcal{T}}(\hat{\vx}),\mathcal{S}_{\theta_\mathcal{S}}(\hat{\vx}))\Big]\;.
    \label{eq:ls_m}
\end{equation}
\section{Experiments}
\subsection{Experimental Settings}
We conducted a comprehensive evaluation of our method across various backbone networks, namely ResNet \cite{resnet}, VGG \cite{vgg}, and WideResNet (WRN)\cite{wrn}, spanning three distinct classification datasets: CIFAR10, CIFAR100 \cite{c10}, and Tiny-ImageNet \cite{tin}. The datasets feature varying scales and complexities, offering a well-rounded assessment of our method's capabilities. In detail, CIFAR10 and CIFAR100 encompass a total of 60,000 images, partitioned into 50,000 for training and 10,000 for testing. CIFAR10 comprises 10 categories, while CIFAR100 boasts 100 categories. The images within both datasets are characterized by a resolution of 32×32 pixels. On the other hand, Tiny-ImageNet comprises 100,000 training images and 10,000 validation images, with a higher resolution of 64 × 64 pixels. This dataset encompasses a diverse array of 200 image categories, contributing to the breadth and comprehensiveness of our evaluation.

\begin{table*}[t]
\caption{The distillation results of compared methods in CIFAR10 and CIFAR100.  The best-performing method is highlighted in bold, and the runner-up is underlined. Additionally, we use superscripts to indicate the sources of these results: $^a$ for \cite{fastdfkd}, $^b$ for \cite{kakr}, $^c$ for \cite{mad}, $^d$ for \cite{spshnet}, and $^e$ for our experiments. In this table, 'R' represents Resnet, 'W' corresponds to WideResnet, and 'V' stands for VGG.}
\centering
\begin{adjustbox}{width=\linewidth}
\begin{tabular}{lcccccccccccc}
\toprule
\textbf{} &
  \multicolumn{5}{c}{\textbf{CIFAR10}} &
  \multicolumn{5}{c}{\textbf{CIFAR100}} &
  \textbf{TinyImageNet} &
  \textbf{ImageNet} \\ \midrule
\multirow{2}{*}{\textbf{Method}} &
  \textbf{R34} &
  \textbf{W402} &
  \textbf{W402} &
  \textbf{W402} &
  \textbf{V11} &
  \textbf{R34} &
  \textbf{W402} &
  \textbf{W402} &
  \textbf{W402} &
  \textbf{V11} &
  \textbf{R34} &
  \textbf{R50} \\
 &
  \textbf{R18} &
  \textbf{W162} &
  \textbf{W161} &
  \textbf{W401} &
  \textbf{R18} &
  \textbf{R18} &
  \textbf{W162} &
  \textbf{W161} &
  \textbf{W401} &
  \textbf{R18} &
  \textbf{R18} &
  \textbf{R50} \\ \midrule
Teacher &
  95.70 &
  94.87 &
  94.87 &
  94.87 &
  92.25 &
  77.94 &
  77.83 &
  75.83 &
  75.83 &
  71.32 &
  66.44 &
  75.45 \\
Student &
  95.20 &
  93.95 &
  91.12 &
  93.94 &
  95.20 &
  77.10 &
  73.56 &
  65.31 &
  72.19 &
  77.10 &
  64.87 &
  75.45 \\
\midrule
DeepInv$^a$ \cite{adi}&
  93.26 &
  89.72 &
  83.04 &
  86.85 &
  90.36 &
  61.32 &
  61.34 &
  53.77 &
  68.58 &
  54.13 &
  - &
  68.00 \\
DFQ$^a$ \cite{dfq}&
  94.61 &
  92.01 &
  86.14 &
  91.69 &
  90.84 &
  77.01 &
  64.79 &
  51.27 &
  54.43 &
  66.21 &
  - &
  - \\
ZSKT$^a$ \cite{zskt}&
  93.32  &
  89.66 &
  83.74 &
  86.07 &
  89.46 &
  67.74 &
  54.59 &
  36.60 &
  53.60 &
  54.31 &
  - &
  - \\
CMI$^a$ \cite{cmi}&
  94.84 &
  92.52 &
  90.01 &
  92.78 &
  91.13 &
  77.04 &
  68.75 &
  57.91 &
  68.88 &
  70.56 &
  64.01 &
  - \\
PREKD$^b$ \cite{predfkd}&
  93.41 &
  - &
  - &
  - &
  - &
  76.93 &
  - &
  - &
  - &
  - &
  49.94 &
  - \\
MBDFKD$^b$ \cite{mbdfkd}&
  93.03 &
  - &
  - &
  - &
  - &
  76.14 &
  - &
  - &
  - &
  - &
  47.96 &
  - \\
FM$^a$ \cite{fastdfkd}&
94.05	&
92.45	&
89.29	&
92.51	&
90.53	&
74.34	&
65.12	&
54.02	&
63.91	&
67.44 &
- &
57.37$^e$ \\
MAD$^c$ \cite{mad}&
  94.90 &
  92.64 &
  - &
  - &
  - &
  77.31 &
  64.05 &
  - &
  - &
  - &
  62.32 &
  - \\
KAKR\_MB$^b$ \cite{kakr}&
  93.73 &
  - &
  - &
  - &
  - &
  77.11 &
  - &
  - &
  - &
  - &
  47.96 &
  - \\
KAKR\_GR$^b$ \cite{kakr}&
  94.02 &
  - &
  - &
  - &
  - &
  77.21 &
  - &
  - &
  - &
  - &
  49.88 &
  - \\
SpaceshipNet$^d$ \cite{spshnet} &
  \textbf{95.39} &
  93.25 &
  90.38 &
  93.56 &
  {\ul 92.27} &
  {\ul 77.41} &
  69.95 &
  58.06 &
  68.78 &
  71.41 &
  {\ul 64.04} &
  - \\ \midrule
\textbf{NAYER ($\mathcal{E} = 100$)} &
  94.03 &
  93.48 &
  91.12 &
  93.57 &
  91.34 &
  76.29 &
  70.20 &
  59.26 &
  69.89 &
  71.10 &
  61.71 &
  - \\
\textbf{NAYER ($\mathcal{E} = 200$)} &
  94.89 &
  {\ul 93.84} &
  {\ul 91.60} &
  {\ul 94.03} &
  91.93 &
  77.07 &
  {\ul 71.22} &
  {\ul 61.90} &
  {\ul 70.68} &
  {\ul 71.53} &
  63.12&
  - \\
\textbf{NAYER ($\mathcal{E} = 300$)} &
  {\ul 95.21} &
  \textbf{94.07} &
  \textbf{91.94} &
  \textbf{94.15} &
  \textbf{92.37} &
  \textbf{77.54} &
  \textbf{71.72} &
  \textbf{62.23} &
  \textbf{71.80} &
  \textbf{71.75} &
  \textbf{64.17} &
  \textbf{68.92} \\ \bottomrule
\end{tabular}
\end{adjustbox}
\label{tab:sota}

\end{table*}

\subsection{Results and Analysis}

\noindent
\textbf{Comparison with SOTA DFKD Methods.} Table \ref{tab:sota} displays the results of DFKD achieved by our methods and several state-of-the-art (SOTA) approaches. In general, previous methods exhibit limitations when generating images from random noise, impacting both training time and image diversity. By using LTE as the input and relocating the source of randomness from the input to the layer level,  our approach provides highly diverse training images and faster running time. Notably, with 300 epochs, our method achieves SOTA performance in all comparison cases, except for the Resnet32/Resnet18 case in CIFAR10. However, it is essential to note that our method was designed in a straightforward manner, without incorporating innovative techniques found in current SOTA approaches, such as activation region constraints and feature exchange in SpaceshipNet  \cite{spshnet}, knowledge acquisition and retention meta-learning in KAKR \cite{kakr}, and momentum distillation in MAD \cite{mad}.

\begin{table*}[]
\caption{Comparing training times in hours using a single NVIDIA A100 for DFKD methods on CIFAR-10 and CIFAR-100 with the teacher/student models WRN40-2/WRN16-2. FM ($\mathcal{E} = 100, 200,$ and $300$) corresponds to the settings of three variants of our methods. We were unable to replicate the training times of KAKR and SpaceshipNet as they did not provide access to their source code.}
\centering
\begin{adjustbox}{width=\linewidth}
\begin{tabular}{@{}lcccccccccccc@{}}
\toprule
 &
  DeepInv &
  CMI &
  DFQ &
  ZSKT &
  MAD &
  SpaceshipNet &
  \multicolumn{1}{c}{\begin{tabular}[c]{@{}c@{}}FM\\ $\mathcal{E}=100$\end{tabular}} &
  \multicolumn{1}{c}{\begin{tabular}[c]{@{}c@{}}FM\\ $\mathcal{E}=200$\end{tabular}} &
  \multicolumn{1}{c}{\begin{tabular}[c]{@{}c@{}}FM\\ $\mathcal{E}=300$\end{tabular}} &
  \multicolumn{1}{c}{\begin{tabular}[c]{@{}c@{}}\textbf{NAYER}\\ $\mathcal{E}=100$\end{tabular}} &
  \multicolumn{1}{c}{\begin{tabular}[c]{@{}c@{}}\textbf{NAYER}\\ $\mathcal{E}=200$\end{tabular}} &
  \multicolumn{1}{c}{\begin{tabular}[c]{@{}c@{}}\textbf{NAYER}\\ $\mathcal{E}=300$\end{tabular}} \\ \midrule
\textbf{CIFAR10} &
  \multicolumn{1}{c}{\begin{tabular}[c]{@{}c@{}}89.72 \\ (31.23h)\end{tabular}} &
  \multicolumn{1}{c}{\begin{tabular}[c]{@{}c@{}}92.52 \\ (24.01h)\end{tabular}} &
  \multicolumn{1}{c}{\begin{tabular}[c]{@{}c@{}}92.01 \\ (3.31h)\end{tabular}} &
  \multicolumn{1}{c}{\begin{tabular}[c]{@{}c@{}}89.66  \\ (3.44h)\end{tabular}} &
  \multicolumn{1}{c}{\begin{tabular}[c]{@{}c@{}}92.64\\ (13.13h)\end{tabular}} &
  \multicolumn{1}{c}{\begin{tabular}[c]{@{}c@{}}93.25\\ (22.35h)\end{tabular}} &
  \multicolumn{1}{c}{\begin{tabular}[c]{@{}c@{}}91.63\\ (2.18h)\end{tabular}} &
  \multicolumn{1}{c}{\begin{tabular}[c]{@{}c@{}}92.05\\ (3.98h)\end{tabular}} &
  \multicolumn{1}{c}{\begin{tabular}[c]{@{}c@{}}92.31\\ (7.02h)\end{tabular}} &
  \multicolumn{1}{c}{\begin{tabular}[c]{@{}c@{}}\textbf{93.48}\\ \textbf{(2.05h)}\end{tabular}} &
  \multicolumn{1}{c}{\begin{tabular}[c]{@{}c@{}}93.84\\ (3.85h)\end{tabular}} &
  \multicolumn{1}{c}{\begin{tabular}[c]{@{}c@{}}94.07\\ (6.78h)\end{tabular}} \\
\textbf{CIFAR100 }&
  \multicolumn{1}{c}{\begin{tabular}[c]{@{}c@{}}61.34\\ (31.23h)\end{tabular}} &
  \multicolumn{1}{c}{\begin{tabular}[c]{@{}c@{}}68.75 \\ (24.01h)\end{tabular}} &
  \multicolumn{1}{c}{\begin{tabular}[c]{@{}c@{}}64.79 \\ (3.31h)\end{tabular}} &
  \multicolumn{1}{c}{\begin{tabular}[c]{@{}c@{}}54.59 \\ (3.44h)\end{tabular}} &
  \multicolumn{1}{c}{\begin{tabular}[c]{@{}c@{}}64.05\\ (26.45h)\end{tabular}} &
  \multicolumn{1}{c}{\begin{tabular}[c]{@{}c@{}}69.95\\ (22.37h)\end{tabular}} &
  \multicolumn{1}{c}{\begin{tabular}[c]{@{}c@{}}67.15\\ (2.23h)\end{tabular}} &
  \multicolumn{1}{c}{\begin{tabular}[c]{@{}c@{}}67.75\\ (4.42h)\end{tabular}} &
  \multicolumn{1}{c}{\begin{tabular}[c]{@{}c@{}}68.25\\ (7.56h)\end{tabular}} &
  \multicolumn{1}{c}{\begin{tabular}[c]{@{}c@{}}\textbf{70.20}\\ \textbf{(2.15h)}\end{tabular}} &
  \multicolumn{1}{c}{\begin{tabular}[c]{@{}c@{}}71.22\\ (4.03h)\end{tabular}} &
  \multicolumn{1}{c}{\begin{tabular}[c]{@{}c@{}}71.72\\ (7.22h)\end{tabular}} \\
\textbf{Avergaing Speed Up} &
  1$\times$ &
  1.3$\times$ &
  9.73$\times$ &
  9.08$\times$ &
  1.78$\times$ &
  1.39$\times$ &
  14.17$\times$ &
  7.46$\times$ &
  4.29$\times$ &
  \textbf{14.88$\times$} &
  7.93$\times$ &
  4.47$\times$ \\ \bottomrule
\end{tabular}
\end{adjustbox}
\label{tab:time}
\end{table*}

\noindent
\textbf{Additional Experiments at Higher Resolution.} To assess the effectiveness of NAYER, we conducted further evaluations on the more challenging ImageNet dataset. ImageNet comprises 1.3 million training images with resolutions of 224×224 pixels, spanning 1,000 categories. ImageNet's complexity surpasses that of CIFAR, making it a significantly more time-consuming task for data-free training. As displayed in Table \ref{tab:sota}, almost all DFKD methods refrain from reporting results on ImageNet due to their prolonged training times. Therefore, our comparison is primarily against DeepInv \cite{adi}, and for the sake of a fair comparison, we re-conducted the experiments of FM \cite{fastdfkd} to align with our settings. The results clearly demonstrate that NAYER outperforms other methods in terms of accuracy, underscoring its efficacy on a large-scale dataset.

\noindent
\textbf{Training Time Comparison.} As shown in Table \ref{tab:time}, the NAYER model trained for 100 epochs (i.e., $\text{NAYER} (\mathcal{E} = 100)$) achieves an average speedup of $15\times$ compared to DeepInv, while also delivering higher accuracies. This substantial speedup is attributed to the significantly fewer steps required for generating samples (30 for CIFAR-10 and 40 for CIFAR-100) compared to DeepInv's 2000 steps. As a result, DeepInv takes over 30 hours to complete training on CIFAR-10/CIFAR-100, whereas our method only requires approximately 2 hours. These results demonstrate that our method not only achieves high accuracy but also significantly accelerates the model training process.

\noindent
\textbf{Additional Experiments in Data-free Quantization.} To demonstrate the use of our data-free generation in other data-free tasks, we further conduct experiments in Data-free Quantization. We conducted a comparative analysis against ZeroQ \cite{zeroq}, DFQ \cite{dfq}, and ZAQ \cite{zaq}. ZeroQ retrains a quantized model using reconstructed data instead of original data, DFQ is a post-training quantization approach that utilizes a weight equalization scheme to eliminate outliers in both weights and activations, and ZAQ is the pioneering method that employs adversarial learning for data-free quantization. In this comparison, our method consistently demonstrated superior accuracy across all four scenarios.

\begin{table}[h]
\centering
\caption{The results of compared methods in Data-free Quantization.}
\begin{adjustbox}{width=\linewidth}
\begin{tabular}{@{}llcccccc@{}}
\toprule
\textbf{Dataset}                   & \textbf{Model}       & \textbf{Bit} & \textbf{Float32} & \textbf{ZeroQ} & \textbf{DFQ} & \textbf{ZAQ} & \textbf{\textbf{NAYER ($\mathcal{E} = 300$)}} \\ 

\midrule
\multirow{2}{*}{\textbf{CIFAR10}}  & \textbf{MobileNetV2} & W6A6         & 92.39            & 89.9           & 85.43        &  {\ul 92.15}        & \textbf{92.23}    \\
                                   & \textbf{VGG19}       & W4A8         & 93.49            & 92.69          & 92.66        &  {\ul 93.06}       & \textbf{93.15}    \\ \midrule
\multirow{2}{*}{\textbf{CIFAR100}} & \textbf{Resnet20}    & W5A5         & 69.58            & 65.7           & 59.42        &  {\ul 67.94}     & \textbf{68.23}    \\
                                   & \textbf{Resnet18}    & W4A4         & 77.38            & 70.25          & 40.35        &  {\ul 72.67}        & \textbf{73.32}    \\ \bottomrule
\end{tabular}
\end{adjustbox}
\label{tab:extin}
\end{table}

\subsection{Ablation Study}
{\color{black}
\noindent
\textbf{Effectiveness of Label-Text Embedding.} We illustrate the impact of using LTE in comparison with random noise (Z) and one-hot vector (OH) as the inputs for the generator. As depicted in first three column in Table \ref{tab:ab_lte_nayer}, LTE demonstrates significantly accelerated averaging convergence in terms of CE Loss. This phenomenon can be attributed to the principle that mapping between two distributions is simplified when they share greater similarity. However, the diversity metric for inputting label information (both LTE and OH) is notably lower than that of random noise. This outcome underscores the adverse effects of the generator overly focusing on constant label information.

\begin{table*}[]
\caption{Comparison with different types of input and random sources involves accuracy, diversity metric and averaging convergence time, which is the average number of epochs the generator needs to synthesize data with Cross-Entropy (CE) Loss $<$ 0.1. Each method undergoes 30 generation steps and runs for 100 epochs. "-" denotes that a model cannot provide any data with CE Loss $<$ 0.1.}
\centering
\begin{adjustbox}{width=1\linewidth}
\begin{tabular}{lccccccccccc}
\hline
                        & OH    & Z     & LTE   & cat    & sum(0.1) & sum(0.5) & sum(1) & NL(woRI)
 & NL(1-to-1) & \textbf{NL(K-to-1)} & NL(woBN) \\ \hline
Averaging Convergence Time(↓) & 28.23    & -     & \textbf{ 8.52}     & 10.47     & 10.17       & 25.12       & -      & 8.68        & 9.67        & 9.53    & 16.58        \\
Diversity   Score(↑)       & 0.013 & 0.137 & 0.016 & 0.0132 & 0.021    & 0.036    & 0.127  & 0.016    & 0.138    & \textbf{0.139}   & 0.131    \\
Accuracy(↑)                & 12.35 & 90.14 & 13.52 & 13.29  & 18.92    & 85.72    & 90.15  & 14.82    & 93.42    & \textbf{93.48}  & 91.15      \\ \hline
\end{tabular}
\end{adjustbox}
\label{tab:ab_lte_nayer}
\end{table*}

\noindent
\textbf{Effectiveness of Noisy Layer.} We analyze the impact of multiple random source strategies, including our NL(1-to-1), NL(K-to-1), NL without reinitiation (WoRI), the concatenation of LTE and random noise Z (cat), and the sum of them (sum($\beta$)): $\vv = \ve_\vy+\beta Z$. Table \ref{tab:ab_lte_nayer} demonstrates that: 1) the sum of LTE and noise have a lower convergence time but higher accuracy and diversity if $\beta$ is high, making them similar to only using random noise $Z$. In contrast, if $\beta$ is low, the convergence time is faster but accuracy and diversity are lower, similar to only using LTE. 2) Using NL boosts the generator's diversity while maintaining rapid convergence and high-quality sampling. 3) Using NL(K-to-1) results in faster convergence and a higher diversity score when compared to using one noisy layer for each individual image (1-to-1). 4) Without reinitiation, the NAYER provides almost similar results to only using LTE, thereby highlighting the effectiveness of reinitiation strategies. The further comparison with different architecture of NL can be found at Supplemental Material.

\noindent
\textbf{Effectiveness of \code{BatchNorm}.} We analyze the impact of \code{BatchNorm} in Eq. \ref{eq:ey}. Table \ref{tab:ab_lte_nayer} shows that without using \code{BatchNorm} (WoBN), NAYER struggles in learning LTE and has lower accuracy (91.15\% compared to 93.48\%), attributed to the close proximity of LTEs, highlighting the effectiveness of using \code{BatchNorm} in our method.

\subsection{Further Analysis}
\label{sec:ana}
\noindent
\textbf{Comparison with Different Memory Size.}
In this comparison, we evaluate the accuracies of our NAYER and MBDFKD models while varying the memory size. Note that, to ensure a fair comparison, we maintain identical generator architectures, including the additional linear layer (noisy layer for NAYER) for MBDFKD. The results demonstrate that: 1) With a bigger memory size, our method can have better performance. 2) Even with only 5k memory size, our method still outperforms the current SOTA DFKD method (90.41\% compared to 90.38\% of SpaceshipNet).

\begin{table}[H]
\centering
\caption{The accuracies of NAYER and MBDFKD with varying the memory buffer size.}
\begin{adjustbox}{width=\linewidth}
\begin{tabular}{@{}lllllllll@{}}
\toprule
\textbf{Memory buffer size} & SOTA & 5k    & 10k   & 20k   & 40k   & 100k  & 200k  & Full  \\ \midrule
MBDFKD            &  90.38 & 73.33 & 74.12 & 73.72 & 72.68 & 71.96 & 71.27 & 70.72\\
NAYER                &  90.38 & \textbf{90.41} & \textbf{90.76} & \textbf{90.98} & \textbf{91.21} & \textbf{91.64} & \textbf{91.86} & \textbf{91.94} \\ \bottomrule
\end{tabular}
\end{adjustbox}
\label{tab:as_mulmem}
\end{table}

\noindent
\textbf{Comparison with Different Prompting Engineering Templates.} We analyze the impact of different prompting engineering techniques to generate the label text. We propose three different ways to prompt the label text, including  P1: \code{"a class of a \{class\_name\}"}, P2: \code{"a photo of a \{class\_name\}"}, P3: \code{"a photo of a \{class\_index\}"}. Table \ref{tab:as_prompt} demonstrates that: 1) P1 and P2 has the best accuracy and far higher than current SOTA; 2) By using only the label index instead of the label name, the performance of P3 remains far better than the best baseline (93.90\% and 71.37\% compared to 93.25\% and 69.95\% for SpaceshipNet). From this, we can infer that using the label index is possible in the datasets with less meaningful labels, further showing the effectiveness of our methods in real-world applications.

\begin{table}[H]
\centering
\caption{Accuracies of different prompt engineering methods.}
\begin{adjustbox}{width=0.9\linewidth}
\begin{tabular}{@{}lcccccccc@{}}
\toprule
             & \multicolumn{4}{c}{CIFAR-10} & \multicolumn{4}{c}{CIFAR-100} \\ \midrule
Text Encoder & SOTA & P1    & P2    & P3  &  SOTA& P1   & P2   & P3    \\
Accuracy           & 93.25 & 93.96 & \textbf{94.07}     & 93.28     & 69.95 & 71.68 & \textbf{71.72}  & 71.12      \\ \bottomrule
\end{tabular}
\end{adjustbox}
\label{tab:as_prompt}
\end{table}

\noindent
\textbf{Comparison with Different Text Encoder.} We evaluate our NAYER using three common text encoders—Doc2Vec \cite{doc2vec}, SBERT \cite{sbert}, and CLIP \cite{clip} (Table \ref{tab:as_multe}). Table \ref{tab:as_multe} shows that: 1) Our method performs well across almost all language models, thanks to their ability to capture the relations of label-text. Furthermore, by combining any language model with the label-index prompt engineering discussed in the previous section, our method is capable of working in various domains, even without meaningful label-text. Secondly, the ability of foundational models like CLIP is demonstrated to improve our model’s performance, attributed to the multimodal nature of its vision-language model. Nonetheless, the improvement is minor (0.09\%). Finally, following the findings, we choose to use CLIP as the text encoder for this paper.

\begin{table}[H]
\centering
\caption{Accuracies of our NAYER with different text encoders.}
\begin{adjustbox}{width=0.9\linewidth}
\begin{tabular}{@{}lcccccccc@{}}
\toprule
             & \multicolumn{4}{c}{CIFAR-10} & \multicolumn{4}{c}{CIFAR-100} \\ \midrule
Text Encoder & SOTA & Doc2Vec    & SBERT    & CLIP  & SOTA & Doc2Vec   & SBERT   & CLIP    \\
Accuracy           & 93.25 & 93.98     & 93.94   & \textbf{94.07}  & 69.95 & 71.58     & 71.63   & \textbf{71.72}   \\ \bottomrule
\end{tabular}
\end{adjustbox}
\label{tab:as_multe}
\end{table}

}

\section{Conclusion}

In this paper, we propose a novel DFKD method, namely NAYER, which utilizes the meaningful LTE as the input and relocates the random source from the input to the noisy layer. The significance of LTE lies in its ability to contain substantial meaningful information, enabling the fast generating images in only few steps. The use of noisy layer address the overfocus problem in using constant input information and increase significantly the diversity. Our extensive experiments on different datasets and tasks prove NAYER's superiority over other SOTA DFKD methods. 

\section*{Acknowledgements}
This work was supported by ARC DP23 grant DP230101176 and by the Air Force Office of Scientific Research under award number FA2386-23-1-4044.

{\small
\bibliographystyle{ieee_fullname}
\bibliography{egbib}
}

\appendix
\section{Training Details}
\subsection{Teacher Model Training Details}

In this work, we employed pretrained ResNet-34 and WideResnet-40-2 teacher models from \cite{fastdfkd} for CIFAR-10 and CIFAR-100. For Tiny ImageNet, we trained ResNet-34 from scratch using PyTorch, and for ImageNet, we utilized the pretrained ResNet-50 from PyTorch. Teacher models were trained with SGD optimizer, initial learning rate of 0.1, momentum of 0.9, and weight decay of 5e-4, using a batch size of 128 for 200 epochs. Learning rate decay followed a cosine annealing schedule.

\begin{table}[h]
\centering
\caption{Generator Network ($\mathcal{G}$) Architecture for CIFAR10, CIFAR100 and TinyImageNet.}
\begin{adjustbox}{width=1\linewidth}
\begin{tabular}{@{}ll@{}}
\toprule
Output          & \textbf{Size   Layers}                              \\ \midrule
1000            & \code{Input    }                             \\
$128 \times h/4 \times w/4$ & \code{Linear, BatchNorm1D, Reshape}                        \\
$128 \times h/4 \times w/4$ & \code{SpectralNorm (Conv (3 × 3)), BatchNorm2D, LeakyReLU} \\
$128 \times h/2 \times w/2$ & \code{UpSample (2×)}                                       \\
$64 \times h/2 \times w/2$  & \code{SpectralNorm (Conv (3 × 3)), BatchNorm2D, LeakyReLU} \\
$64 \times h \times w$     & \code{UpSample (2×)  }                                     \\
$3 \times h \times w$      & \code{SpectralNorm (Conv (3 × 3)), Sigmoid, BatchNorm2D  }    \\ \bottomrule
\end{tabular}
\end{adjustbox}
\label{tab:g_arch}
\end{table}
\begin{table}[h]
\centering
\caption{Generator Network ($\mathcal{G}$) Architecture for ImageNet.}
\begin{adjustbox}{width=1\linewidth}
\begin{tabular}{@{}ll@{}}
\toprule
Output          & \textbf{Size   Layers}                              \\ \midrule
1000            & \code{Input}                                 \\
$128 \times h/16 \times w/16$ & \code{Linear, BatchNorm1D, Reshape  }                      \\
$128 \times h/16 \times w/16$ & \code{SpectralNorm (Conv (3 × 3)), BatchNorm2D, LeakyReLU} \\
$128 \times h/8 \times w/8$ & \code{UpSample (2×)} \\                    
$128 \times h/8 \times w/8$ & \code{SpectralNorm (Conv (3 × 3)), BatchNorm2D, LeakyReLU} \\
$128 \times h/4 \times w/4$ & \code{UpSample (2×)} \\                      
$64 \times h/4 \times w/4$ & \code{SpectralNorm (Conv (3 × 3)), BatchNorm2D, LeakyReLU} \\
$64 \times h/2 \times w/2$ & \code{UpSample (2×)} \\
$64 \times h/2 \times w/2$  & \code{SpectralNorm (Conv (3 × 3)), BatchNorm2D, LeakyReLU} \\
$64 \times h \times w$     & \code{UpSample (2×)}                                       \\
$3 \times h \times w$      & \code{SpectralNorm (Conv (3 × 3)), Sigmoid, BatchNorm2D}      \\ \bottomrule
\end{tabular}
\end{adjustbox}
\label{tab:g_arch_in}
\end{table}
\begin{table}[h]
\caption{The hyperparameters for NAYER applied to four different datasets are detailed below. Specifically, $\alpha_{cls}$, $\alpha_{bn}$, and $\alpha_{adv}$ are the hyperparameters associated with Eq.~(\ref{eq:lz}), and their values are consistent with the settings defined in \cite{fastdfkd}. The variables $I$ and $S$ denote the number of iterations for generating and training the student, respectively, while $g$ represents the training steps to optimize the generator $\mathcal{G}_{\theta_{\mathcal{G}}}$ and the noisy layers $\mathcal{Z}$.}
\begin{adjustbox}{width=1\linewidth}
\begin{tabular}{@{}lllllllll@{}}
\toprule
             & batch size (student) & batch size (generator) & $\alpha_{cls}$ & $\alpha_{bn}$ & $\alpha_{adv}$  & $I$  & $g$   & $S$    \\ \midrule
CIFAR10      & 512                  & 400                    & 0.5 & 10 & 1.33 & 2 & 30  & 400     \\
CIFAR100     & 512                  & 400                    & 0.5 & 10 & 1.33 & 2 & 40  & 400   \\
TinyImageNet & 256                  & 200                    & 0.5 & 10 & 1.33 & 4 & 60  & 1000  \\
ImageNet     & 128                  & 50                     & 0.1 & 0.1  & 0.1 & 20 & 100 & 2000 \\ \bottomrule
\end{tabular}
\end{adjustbox}
\label{tab:hyperpara}
\end{table}

\subsection{Student Model Training Details}

To ensure fair comparisons, we adopt the generator architecture outlined in \cite{fastdfkd} for all experiments. Specifically, the generator architecture for CIFAR10, CIFAR100, and TinyImageNet is elaborated upon in Table \ref{tab:g_arch}, while the generator architecture for ImageNet is provided in Table \ref{tab:g_arch_in}. Across all experiments, we maintain a consistent approach for training the student model, employing a batch size of 512. We utilize the SGD optimizer with a momentum of 0.9 and a variable learning rate, following a cosine annealing schedule that starts at 0.1 and ends at 0, to optimize the student parameters ($\theta_\mathcal{S}$). Additionally, we employ the Adam optimizer with a learning rate of 4e-3 for optimizing the generator.We present the results in three distinct variants, each corresponding to a different value of $\mathcal{E}$: 100, 200, and 300, all incorporating a configuration of 20 warm-up epochs, in line with the settings defined in \cite{fastdfkd}. Further details regarding the parameters can be found in Table \ref{tab:hyperpara}.

\section{Extended Results}
\subsection{{Experiments in Segmentation:}} In response to your feedback, we conducted semantic segmentation experiments following FM [1] settings. By utilizing dataset part names such as 'Basements, Bathrooms, ...' for LTE and the Noisy Layer as the random source, our method in Table \ref{tab:seg} outperforms previous works with better IoU.
\begin{table}[h]
\centering
\begin{adjustbox}{width=0.9\linewidth}
\begin{tabular}{@{}lcccc@{}}
\toprule
Method    & DFAD  & DAFL  & FM    & NAYER \\ \midrule
Synthetic Time & 6.0h  & 3.99h & \textbf{0.82h} & \textbf{0.82h} \\
mIoU      & 0.364 & 0.105 & 0.366 & \textbf{0.385} \\ \bottomrule
\end{tabular}
\end{adjustbox}
\caption{Mean IoU on NYUv2 Segmentation dataset.}
\label{tab:seg}
\end{table}

\subsection{Noisy Layer Architecture}
In Table \ref{tab:nlarch}, we compare the different architectures in terms of: 
\begin{itemize}[noitemsep,nolistsep]
    \item The averaging accuracy.
    \item The averaging convergence time, which is the average number of epochs the generator needs to synthesize data with Cross-Entropy (CE) Loss $<$ 0.1.
    \item The averaging diversity metric, which is calculated using the average L2 distance between the features of new and old data.
\end{itemize}
The different architectures for the Noisy Layer include:
\begin{itemize}[noitemsep,nolistsep]
    \item A1: \code{Linear}
    \item A2: \code{Linear}, \code{Linear}
    \item A3: \code{BatchNorm}, \code{Linear}
    \item A4: \code{BatchNorm}, \code{Linear}, \code{Linear}
    \item A5: \code{BatchNorm}, \code{Linear}, \code{Sigmoid}
    \item A6: \code{BatchNorm}, \code{Linear}, \code{Tanh}
    \item A7: \code{BatchNorm}, \code{Linear}, \code{ReLU}
    \item A8: \code{BatchNorm}, \code{Linear}, \code{Dropout}
\end{itemize}
The result demonstrates that: 
\begin{itemize}[noitemsep,nolistsep]
    \item The combination of \code{BatchNorm} and single \code{Linear} layer produce the best performance.
    \item The architecture of multi \code{Linear} layerslayers results in a longer convergence time and subsequently slightly reduces accuracy. 
    \item The \code{BatchNorm} pplays an important role in improving accuracy, reducing convergence time by increasing the difference between LTEs.
    \item The activation function such as \code{ReLU, Sigmoid} and \code{Tanh} do not improve the performance of our NAYER.
\end{itemize}

\begin{table}[H]
\caption{The accuracies of our NAYER and FM (which uses random noise as the input) with varying training steps for generators.}
\begin{adjustbox}{width=\linewidth}
\begin{tabular}{@{}lcccccccc@{}}
\toprule
                        & A1    & A2    & A3             & A4    & A5    & A6    & A7    & A8    \\ \midrule
Avg.   Convergence Time & 16.58 & 22.72 & 9.53           & 15.73 & 9.63  & 9.59  & 9.61  & 12.72 \\
Diversity   Score       & 0.131 & 0.137 & 0.139          & 0.141 & 0.137 & 0.135 & 0.138 & 0.138 \\
Accuracy                & 92.25 & 92.11 & \textbf{93.48} & 93.37 & 93.41 & 93.37 & 93.42 & 93.42 \\ \bottomrule
\end{tabular}
\end{adjustbox}
\label{tab:nlarch}
\end{table}

\subsection{Comparison with Different Generation Steps}

We compare NAYER and FM, both utilizing random noise as input, while adjusting the training steps for their generators. It's important to note that for a fair comparison, we employ the same generator architectures, including the additional linear layer (noisy layer for NAYER) for FM. Furthermore, all models are trained for 300 epochs. This approach allows us to assess their performance under consistent conditions and understand how varying the generator training steps impact their accuracy. The results indicate that our method has the best results with 40 generation steps for CIFAR100 and 30 steps for CIFAR10. Furthermore, NAYER outperforms FM in all cases of generator training steps.

\begin{table}[h]
\centering
\caption{The accuracies of our NAYER and FM (which uses random noise as the input) with varying training steps for generators.}
\begin{adjustbox}{width=\linewidth}
\begin{tabular}{@{}lccccccc@{}}
\toprule
\textbf{Generator's training steps  }  & $g = 2$ & $g = 5$ & $g = 10$ & $g = 20$ & $g = 30$ & $g = 40$ & $g = 50$ \\ \midrule
\textbf{FM} & 57.08 & 63.83 & 65.12  & 66.82  & 67.51  & 68.23 & 68.18  \\
\textbf{NAYER} & \textbf{59.23} & \textbf{65.14} &\textbf{ 68.13}  & \textbf{69.31}  & \textbf{70.42}  & \textbf{71.72}  & \textbf{71.70}   \\ \bottomrule
\end{tabular}
\end{adjustbox}
\label{tab:as_mulfm}
\end{table}

\subsection{{Robust experiments}} Thanks for your comments. The robust experiments in three runs in Table \ref{tab:robust} shows our method's consistently better accuracy with only minor standard deviation. Notably, previous works omitted these numbers, and due to their high complexity, we did not replicate their results in this rebuttal period. 
\begin{table}[h]
\centering
\caption{Averaging accuracy and standard deviation in three runs.}
\begin{adjustbox}{width=\linewidth}
\begin{tabular}{@{}lcccccc@{}}
\toprule
             & \multicolumn{3}{c}{CIFAR10}                      & \multicolumn{3}{c}{CIFAR100}                     \\ \midrule
            & R34/R18        & W402/W162      & W402/W161      & R34/R18        & W402/W162      & W402/W161      \\ 
SpaceshipNet & \textbf{95.39} & 93.25          & 90.38          & 77.41        & 69.95         & 58.06         \\
NAYER ($\mathcal{E}=300$)     & 95.24 $\pm$ 0.15          & \textbf{94.11 $\pm$ 0.18} & \textbf{91.94 $\pm$ 0.15} & \textbf{77.56 $\pm$ 0.12} & \textbf{71.72 $\pm$ 0.14} & \textbf{62.23 $\pm$ 0.21} \\ \bottomrule
\end{tabular}
\end{adjustbox}
\label{tab:robust}
\end{table}

\subsection{NAYER without Label Text Embedding (LTE)} To highlight our method's LTE-independent capability, we conducted experiments using one-hot vectors and Noisy Layer in Table \ref{tab:woLTE}. Despite the lower accuracy of the one-hot version compared to the LTE version, our method still outperforms the SOTA approach in both scenarios.

\begin{table}[h]
\centering
\caption{Accuracy in CIFAR10 with W402/W162 Architecture.}
\begin{adjustbox}{width=\linewidth}
\begin{tabular}{@{}lccc@{}}
\toprule
Method   & SpaceshipNet & NAYER with LTE & NAYER with one-hot vector \\ \midrule
CIFAR10  & 93.25        & 94.07 & 93.72           \\
CIFAR100 & 69.95        & 71.72 & 70.78           \\ \bottomrule
\end{tabular}
\end{adjustbox}
\label{tab:woLTE}
\end{table}

\subsection{{Non-BatchNorm Architecture}}  The need for batch norm loss is the limitation for most SOTA DFKD methods. Therefore, exploring high performance with non-batchnorm architectures is an intriguing future direction. For this rebuttal, we conduct the CIFAR10 experiments with AlexNet as the student (Table \ref{tab:alex}). The results suggest that our NAYER outperforms previous work when applied to AlexNet.

\begin{table}[h]
\centering
\caption{Accuracy with AlexNet Student.}
\begin{adjustbox}{width=\linewidth}
\begin{tabular}{@{}lcccc@{}}
\toprule
         & Teacher Accuracy & Student Accuracy & FM ($\mathcal{E} = 300$)     & NAYER ($\mathcal{E} = 300$)  \\ \midrule
AlexNet/AlexNet  & 74.74\% & 74.74\% & 65.37\% & 70.14\% \\
Resnet34/AlexNet & 95.70\% & 74.74\% & 68.38\% & 71.15\%        \\ \bottomrule
\end{tabular}
\end{adjustbox}
\label{tab:alex}
\end{table}

\subsection{{Additional Abalation Studies for Noisy Layer}} Inspired by your recommendations, we conducted additional experiments in Table \ref{tab:ab}. The results show that: (4.1) NL with reinitialization (wRI) outperforms with out reinitialization (woRI); (4.2) With beta greater than one, the sum method performs worse than the NL; (4.3) While we used normal noise for the sum method, we further experimented with uniform noise (uni); however, the results remained significantly lower than our NL; (4.4) We will include the note of a zero bias in our revised paper.

\begin{table}[h]
\centering
\caption{Additional ablation Study for Noisy Layer}
\begin{adjustbox}{width=\linewidth}
\begin{tabular}{@{}l|cc|ccc|ccc@{}}
\toprule
Method                        & NL(woRI) & NL(wRI) & sum(1.5) & sum(2.0) & sum(3.0) & uni(1.0) & uni(1.5) & uni(2.0) \\ \midrule
Avg. Convergence Time(↓) & 8.68     & 9.53    & -        & -        & -        & -            & -            & -            \\
Diversity Score(↑)            & 0.016    & 0.139   & 0.129    & 0.133    & 0.135    & 0.113        & 0.128        & 0.139        \\
Accuracy(↑)                   & 14.82    & \textbf{93.48}   & 90.23    & 90.15    & 89.12    & 87.75        & 88.21        & 87.15        \\ \bottomrule
\end{tabular}
\end{adjustbox}
\label{tab:ab}
\end{table}

\subsection{t-SNE Visuallization of LTE and Ground-truth Dataset Distribution}

In this section, we aim to illustrate the interclass information captured by LTE (Label-Text Embedding). To achieve this, we provide t-SNE visualizations of the embeddings for labels and ground-truth data distribution pertaining to four distinct classes: Car, Cat, Dog, and Truck. The t-SNE representation of LTE closely aligns with the ground-truth distribution, especially in the proximity between classes like Car and Truck, as well as Cat and Dog, indicating notably smaller distances compared to other class pairings.

\begin{figure}[t]
\begin{center}
\includegraphics[width=\linewidth]{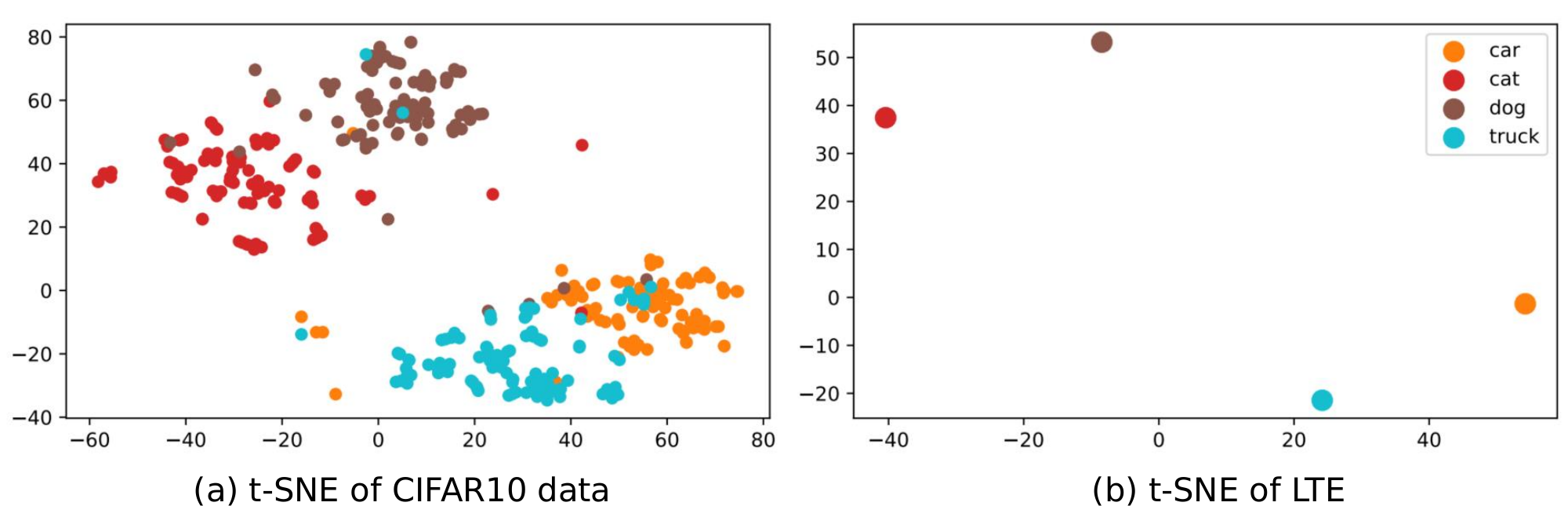}
\end{center}
\caption{t-SNE Visualization of Label-Text Embedding and Ground-Truth Dataset Distribution for Four Classes: Car, Cat, Dog, and Truck.}
\label{fig:tsne}
\end{figure}

\subsection{Visualization.} 
\begin{figure*}[t]
\begin{center}
\includegraphics[width=1\linewidth]{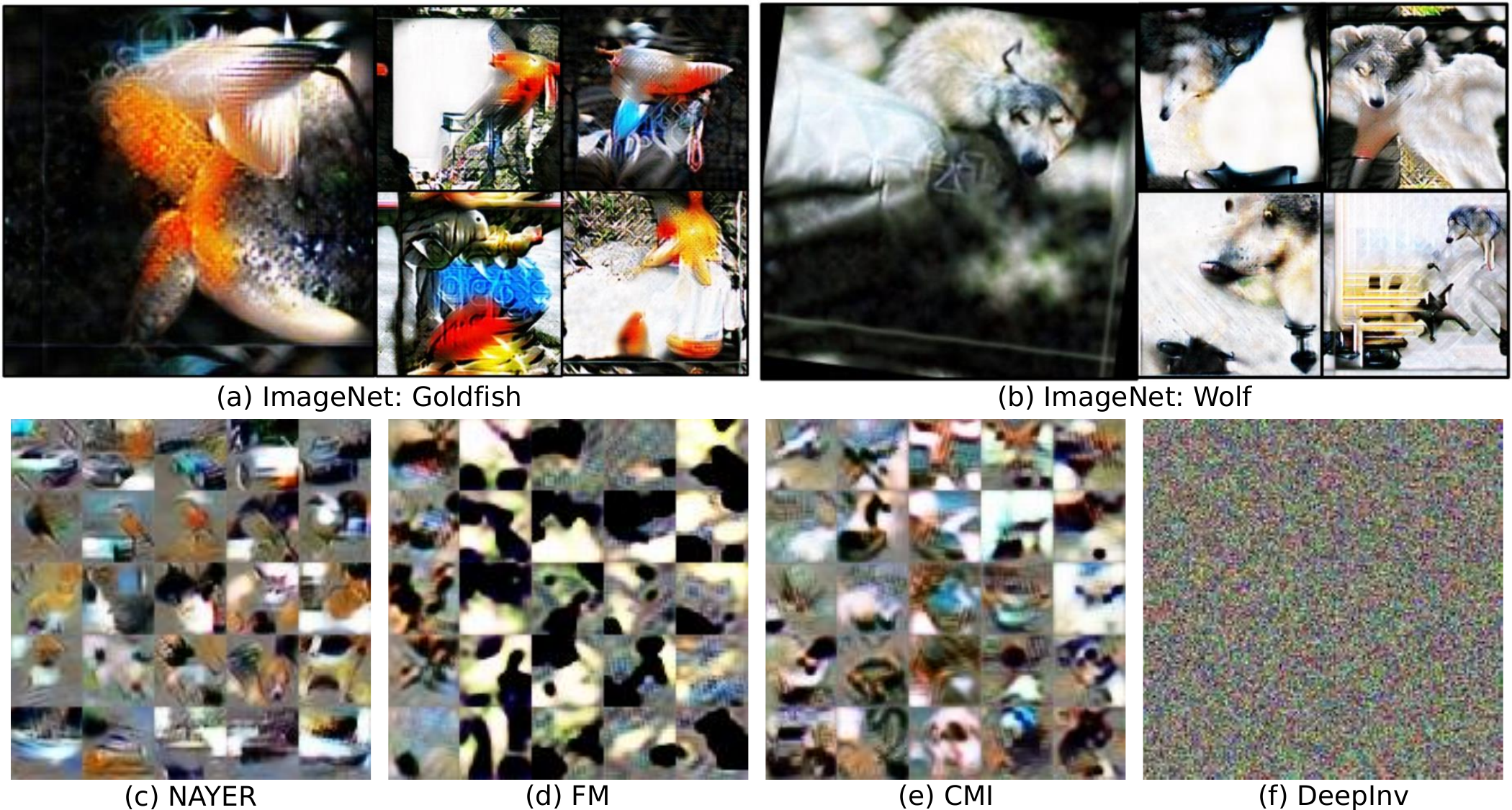}
\end{center}
\caption{(a, b) Display synthetic data generated by our NAYER for ImageNet in just 100 steps. (c, d, e, f) Showcase synthetic data generated for 5 classes (from top to bottom: Car, Bird, Cat, Dog, Ship) in CIFAR10, using only 20 steps of NAYER, FM, CMI, and DeepInv.}
\label{fig:visual}
\end{figure*}

The synthetic results achieved by NAYER within just 100 generator training steps on ImageNet by employing the ResNet-50 as teacher model are presented in Figure~\ref{fig:visual}a-b. For further comparison, we also visualize synthetic images generated by NAYER, FM, CMI, and DeepInv in Figure~\ref{fig:visual}c-f. All of these samples are generated using 20 steps with a ResNet-34 teacher model in the CIFAR-10 dataset. While it remains challenging for human recognition and significantly differs from real datasets, our synthetic images contain common knowledge that represents the classes, thereby visibly demonstrating superior quality compared to other methods.

\section{Further Discussion}

\noindent
\textbf{Does NAYER Preserve a Data-free Setting?} In the definition, data-free knowledge distillation is characterized by training a model without direct access to the teacher model's training data, stemming from the necessity for privacy in datasets. Therefore, the introduction of static label text embeddings does not violate this definition as it does not interact with any training data. In real-world applications, obtaining label embeddings from publicly available pre-trained language models like CLIP or ChatGPT is quick and straightforward. Crucially, our method requires no fine-tuning or retraining of these pretrained models, and thus it does not use any external data. Consequently, it can seamlessly adapt to any real-world DFKD application.


\noindent
\textbf{Does NAYER Work in Meaningless Label Datasets?} The paper suggests the utilization of label-text embedding (LTE), acknowledging its potential drawbacks in datasets lacking meaningful labels, such as those involving chemical compounds. However, our approach highlights two key advantages associated with LTE. Firstly, LTE acts as a dense vector, encompassing more information and thereby facilitating a smoother learning process for the model. Secondly, LTE has the ability to depict relationships between classes. In cases where a class lacks a meaningful label, we can still utilize the class index to generate a label text, such as \code{"a class of \{class\_index\}"}, enabling the extraction of LTE.

While this approach may not completely capture the relationships between classes, it provides richer information compared to a one-hot vector. To support this assertion, we conducted an ablation study on various prompt engineering techniques in Section 4.4. The results indicate that when using only the label index instead of the label name, the performance of P3 remains significantly superior to the best baseline. Specifically, this template P3 achieves 93.72\% accuracy compared to SpaceshipNet (the state-of-the-art model) with 93.25\%, and 71.17\% compared to 69.95\%, respectively. This underscores the viability of employing the label index, particularly in datasets with less meaningful labels, further validating the effectiveness of our methods in real-world applications.

\section{Furture Works}

The proposed NAYER does not incorporate the innovative techniques utilized in current SOTA methods, such as feature mixup \cite{spshnet}, knowledge acquisition and retention \cite{kakr}, and momentum updating \cite{mad}. This leaves space for potential improvements through the integration of these techniques in the future. Additionally, NAYER can be applied to various data-free methods, including but not limited to data-free quantization or data-free model stealing.
\end{document}